\documentclass[runningheads]{llncs}
\usepackage[T1]{fontenc}
\usepackage{graphicx}
\begin{document}
\title{Trashbusters: Deep Learning Approach for Litter Detection and Tracking}
%
%
\author{Kashish Jain\inst{1}, Manthan Juthani\inst{2}, Jash Jain\inst{3}, \and
Anant V. Nimkar\inst{4}}
%
%
\institute{Sardar Patel Institute Of Technology, Mumbai, India \and
\email{\{kashish.jain, manthan.juthani, jash.jain, anant\_nimkar\}@spit.ac.in}}

\maketitle              
%

%
%
%

\begin{abstract}
The illegal disposal of trash is a major public health and environmental concern. Disposing of trash in unplanned places poses serious health and environmental risks. We should try to restrict public trash cans as much as possible. This research focuses on automating the penalization of litterbugs, addressing the persistent problem of littering in public places. Traditional approaches relying on manual intervention and witness reporting suffer from delays, inaccuracies, and anonymity issues. To overcome these challenges, this paper proposes a fully automated system that utilizes surveillance cameras and advanced computer vision algorithms for litter detection, object tracking, and face recognition. The system accurately identifies and tracks individuals engaged in littering activities, attaches their identities through face recognition, and enables efficient enforcement of anti-littering policies. By reducing reliance on manual intervention, minimizing human error, and providing prompt identification, the proposed system offers significant advantages in addressing littering incidents. The primary contribution of this research lies in the implementation of the proposed system, leveraging advanced technologies to enhance surveillance operations and automate the penalization of litterbugs.
\end{abstract}


\vspace{0.5cm}
\section{Introduction}

The spread of surveillance equipment has substantially improved public space monitoring and safety. Public littering requires manual surveillance to catch offenders. The study tries to automate litterbug punishment.

Traditional responses to littering issues focus mainly on manual intervention and witness reporting, which slows response and may misidentify perpetrators. There are further litterbug identification concerns that affect responsibility. Traditional methods are too resource-intensive to solve the challenge. Only automation and surveillance cameras will work. This project attempts to fix the problem by catching the arbitrator polluting and then utilising identification cards in the database to identify the litterbug and penalise them without user intervention..

There is no fully automated method for punishing public littering. However, surveillance camera detection research laid the groundwork for this paper. Using deep learning techniques, Córdova et al. \cite{b1} suggested a real-time littering detection system for smart cities. This method was offered but not implemented. The authorities must still identify and punish the litterbug. This study takes inspiration from their work by implementing the solution and improving their litter detection and tracking algorithms by adding litterbug identification and penalising to make it fully automated.

The solution tracks littering with public surveillance cameras. Advanced computer vision techniques like YOLOv4 and Area Over Intersection correctly detect litterers. The DeepSORT system tracks objects reliably, while face recognition records litterers' distant faces. These photos are compared to a database to accurately identify culprits for punishment. This integrated technology quickly identifies and responds to littering instances, enforcing anti-littering policies. Automation lowers human mistake and subjective judgement, improving reliability.

Implementing the litterbug detection and identification system is the main contribution. We want to reduce manual intervention and improve surveillance operations by leveraging and improving modern computer vision and artificial intelligence technology. The system's revolutionary combination of automated litter detection, object tracking, and face recognition technology improves identification and enforcement.

The article presents a systematic approach to addressing littering in public settings, which is detailed throughout the document.
The second portion analyses recent advances in computer vision, object tracking, and facial recognition technology. Section III explains how the proposed system differs from others through research contributions. Section IV outlines and discusses the evaluation procedure and instruments. In Section V, findings and interpretations are explored.
The key findings and future recommendations are presented in Section VI.

\section{Literature Survey}

An examination of the available literature on systems for the identification and tracking of litter reveals that great headway has been made in this area of research and development. However, there are still a number of issues that need to be addressed, such as recognizing faces from afar and monitoring various objects in scenarios with a lot of people.

Regarding the detection of litter, Korchani et al. \cite{b1}  proposed a real-time littering detection system that makes use of deep learning techniques. Their system is made up of two different modules: one is a litter detection module that uses YOLO-v3 to identify objects that people have thrown, and the other is a person tracking module that makes use of SORT to identify the person who threw the object and follow them until the camera is able to photograph their face. Yun et al. \cite{abcd1} provided an innovative methodology for identifying actions involving the dumping of garbage in real-world surveillance footage. The accuracy of their system was measured at 92 percent when it was applied to a dataset of real-world surveillance videos. Nguyen et al. \cite{defg} presented an innovative data-driven method for detecting change and motion in video analysis and surveillance systems. Their method combines a background model that is based on samples with a feature extractor that is developed through the training of a triplet network.

Regarding the topic of face recognition, Babu et al. \cite{fr} concentrate on automated facial recognition (AFR) in real-world environments. They talk about the difficulties and restrictions that are related with face recognition. They investigate previously developed methods such as cascade classifiers and highlight the performance-enhancing potential of convolutional neural networks (CNNs). Yao et al. \cite{frad2} make use of their UTK-LRHM face database in order to conduct research on face recognition at large distances. 

Dang et al. \cite{ds} propose an enhanced architecture for object tracking by combining YOLOv3 and the Deep SORT tracking algorithm. Their architecture aims to improve multiple object tracking speed and accuracy. Although the paper lacks detailed explanations of the enhancements and comparative evaluations, it serves as a baseline for our use of DeepSORT and provides a starting point

In their work, Cenn et al. \cite{b3} analyzed the reasons, consequences, and possible solutions for littering. They identified factors influencing littering behavior, such as personal attitudes, social norms, situational factors, and perceived consequences. The study also discussed the environmental, social, and economic impacts of littering, along with proposed solutions including education, awareness campaigns, enforcement, and incentives. While this work provides valuable insights into the problem of littering, it does not focus on the technical aspects of detection or tracking.

These publications provide a complete review of current trash detection and tracking technologies, as well as a summary of their findings. They highlight the constraints and limitations of existing approaches and provide insights into potential future avenues for study in this field. In addition, they highlight the issues and limitations of existing methodologies.

\section{Trashbusters}

A concurrent active tracking system for face detection and an automated garbage dumping activity detection system will develop an automatic penalizing system. We proposed using CCTVs to catch road litterers. In the following frames, we use DeepSORT and FRAD to capture the litterbug's face. We match facial detection findings to our person database and punish the person on their national identity card. \par
Our main goal was to identify pedestrian littering from other cleanup efforts. We employed various object detection models with state-of-the-art findings for this. Once the cleaning action is identified, a reverse strategy is used to detect it simultaneously. \par

\subsection{Littering Detection}

We tried separating a person as a pixelated blob to identify tossing or littering. If the human became a pixelated binary blob, we may see another moving away. The second glob was trash. This suggested scene-change videos. This method yields inconsistent results due to photo fragmentation.  \par
The You Only Look Once (YOLO) system \cite{emu1} recognises objects in videos, live streams, and pictures. The YOLO algorithm detects things using deep convolutional neural network characteristics. Its prediction uses eleven distinct convolutions, verifying the feature map and prediction map have the same size. Thirteen of eighty Yolo item classes were litter. Bottles, bags, umbrellas, wallets, bananas, and apples were included. \par

Humans in the stream were recognised using HOG. The HOG detector uses a sliding window to recognise individuals across the image.The adoption of HOG, a texture-based detection method, was justified by its advantages and features as described in [12]. Each location in the detection window receives a HOG descriptor. The trained SVM analyses the description and identifies it as "a person" or "not a person." The bounding box borders were extended by a constant to minimise the number of boxes around an individual to one.

The Yolo Model identified waste, then the AOI approach was used. Another axis-aligned bounding box always results from two overlapping ones. Overlap was used to calculate where the person and trash intersect. Once the rubbish is off the person, the area is zero and the culprit may be recognised.\par

On the other side, cleaning is accomplished when trash reaches a person's bounding box and shifts up or down. Our model was verified using a webcam and a previously recorded video feed. You could find people cleaning up or throwing rubbish using either way. \par

\subsection{Improved DeepSORT}

Our upgraded version of DeepSORT, a traditional multi-object tracking approach (MOT17), is based on \cite{deepyolo}. It has detection, feature extraction, trajectory association, and post-processing modules. A cutting-edge object detector (YOLO) and a feature embedding model (BoT or simpleCNN) extract bounding boxes and appearance features from each frame in the detection and feature extraction module1. The trajectory association module uses the Unscented Kalman filter to predict tracklet motion and the Hungarian algorithm to assign detections based on a motion and appearance cost matrix1. The post-processing module fixes MOT's missing association and detection issues with AFLink and GSI. AFLink widely associates short tracklets without appearance. RNNs encode tracklet motion patterns, while GNNs learn tracklet affinity from temporal overlap and motion similarity1. Gaussian-smoothed interpolation (GSI) estimates object locations from earlier and subsequent observations using Gaussian process regression to account for missed detections. We resolved occlusion, camera movement, and view point changes.

\begin{figure}[h]
\centering

\includegraphics[width=10cm]{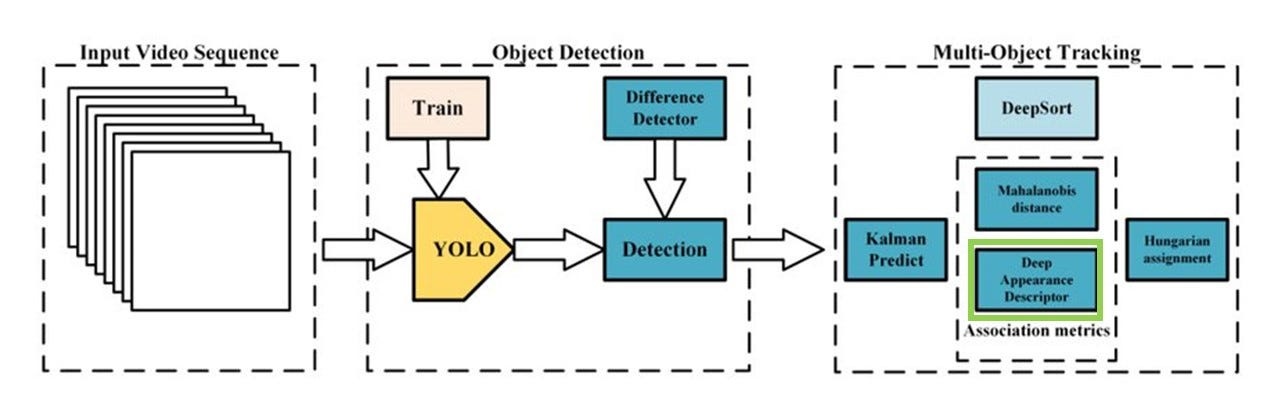}
\caption{DeepSORT}

\includegraphics[width=10cm]{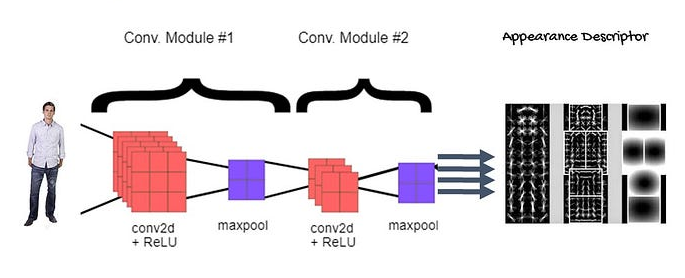}
\caption{DeepSORT Appearance Descriptor}
\end{figure}

The Unscented Kalman filter predicts the motion state of each tracklet as:
\begin{equation}
\label{eq:mgf}
\hat{x}_k = F_k x_{k-1} + B_k u_k
\end{equation}
here $\hat{x}_k$ is the predicted state vector, $F_k$ is the transition matrix, $x_k1$ is the previous state vector, $B_k$ is the control input matrix, and $u_k$ is the control vector.
\newline\newline
The Hungarian algorithm assigns the detections to the tracklets by minimizing a cost matrix that is defined as:
\begin{equation}
\label{eq:mgf}
C_{ij} = \lambda_m d_m(i,j) + \lambda_a d_a(i,j)
\end{equation}
where $C_{ij}$ is the price of assigning detection i to tracklet j , $\lambda_m$ and $\lambda_a$ are weighting factors, and $d_m(i,j)$ and $d_a(i,j)$ are the motion and appearance distances between detection i and tracklet j.
\newline\newline
The Gaussian process regression estimates the location of missing detections as:
\begin{equation}
\label{eq:mgf}
y_* = K(x_*, X)(K(X,X) + \sigma_n^2 I)^{-1} y
\end{equation}
where $y_*$ is the predicted location vector, $x_*$ is the query vector (containing frame index), X is the training input matrix (containing frame indices of previous and subsequent observations), y is the training output matrix (containing locations of previous and subsequent observations), K() is a kernel function (such as RBF), and $\sigma_n^2 I$ is a noise term.

\subsection{FRAD (Face Recognition at a Distance)}
We employed multi-task CNN to improve face detection at long distances, scales, and light conditions for FRAD. We coupled this with Arcface, a cutting-edge face recognition system. MTCNN and Arcface can successfully detect and distinguish faces in real-world scenarios, even in difficult conditions. This method improves face detection at long distances, varied sizes, and different lighting situations while ensuring accurate and reliable face recognition based on learnt feature representations.

\subsubsection{Multi-task CNN}:
Proposal Network (P-Net), Refine Network (R-Net), and Output Network (O-Net) are the three convolutional neural networks (CNNs) that make up the Multi-Task Cascaded Neural Network (MTCNN).

\begin{enumerate}
\item P-Net: A shallow CNN quickly creates candidate windows. The network returns facial landmark coordinates and the image's bounding box from an input image. After that, bounding box regression refines the boxes.

\item R-Net: The P-Net-generated windows are filtered again to eliminate most erroneous positives. The remaining windows are squared and passed to a second CNN to increase bounding boxes and facial landmark placements.

\item O-Net: The third phase is similar the second but employs a more advanced CNN. The face classification scores, final bounding box, and facial landmark placements are now available.
\end{enumerate}
The loss function for MTCNN is a multi-task loss that combines classification loss, bounding box regression loss, and landmark regression loss.

\subsubsection{Additive Angular Margin Loss(ArcFace)}:
For each recognized face, bounding box coordinates clip a region of interest (ROI) from the original image. The Arcface model generates the feature vector from the clipped ROI. Arcface maps face pictures to learn discriminative features by using cosine similarity to put faces of the same person together and push faces of others apart in a high-dimensional feature space. 

Given a feature vector x and its corresponding weight vector W, the cosine similarity between them is defined as: 
\begin{equation}
\label{eq:mgf}
cos(\theta) = (W^T * x) / (|W| * |x|)
\end{equation}
ArcFace adds an additive margin m to the decision boundary by altering the goal logit to cos($\theta + m$). This strategy improves intra-class compactness and inter-class discrepancy by angularly separating the ground truth class from the other classes.

\section{Experimental Setup}

The state-of-the-art YOLOv4 model was utilized to improve litter detection in the study. Initially, spotting bottles, handbags, and umbrellas was difficult. To circumvent this restriction, the YOLOv4 model was fine-tuned using a bespoke dataset, as described in \cite{yolo}. Cropping and rotation were used to segment a variety of litter photos. The model's litter prediction accuracy improved after 100 epochs of 0.001 learning.

\begin{figure}[h]
\centering
\includegraphics[width=8cm]{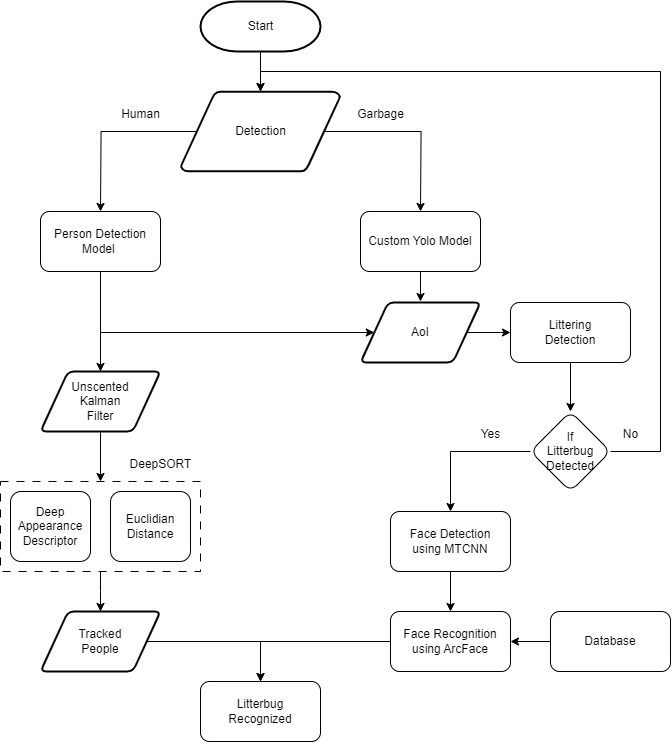}
\caption{Architectural Diagram}
\end{figure}

The research's problem statement lists two types of challenges that typical tracking algorithms cannot tackle. First, occlusion. Occlusion makes it difficult for algorithms to track items that are partially or totally obscured by other objects or impediments. Second, as the camera or observer changes viewpoint, the monitored object's appearance and shape can vary, causing tracking mistakes and mismatches. These algorithms cannot consistently and accurately track objects across diverse viewpoints, limiting their application in complicated real-world contexts. These difficulties were appropriately addressed by the DeepSORT approach. Traditional DeepSORT employs the Kalman Filter, which assumes linearity and ignores non-linear dynamics like people's movements. The study challenge involves non-linear motion, thus we had to incorporate a novel technique to DeepSORT. As mentioned in \cite{ukf}, the DeepSORT algorithm was integrated with the Unscented Kalman Filter (UKF). The UKF helped researchers approximate non-linear transformations using the Unscented Transform. This improved state variable estimates for monitored objects, including position and velocity.

The study team replicated a real-life scenario to strengthen the system. They tried to capture a person's face from identification cards instead of taking a picture. This involved collecting Aadhar and college ID cards from many people. A technique was created to extract and recognise faces from ID cards to map littering by face. This offers perfect automation since Trashbuster users only need to add their ID card to the database.
Webcam video streams were fed into the person detection model to integrate the experimental setup. The tracking Deep Sort model assigned unique IDs to tracked individuals using the discovered person traits. At the same time, the researchers used MTCNN for face detection and ArcFace for face recognition. The researchers labelled each participant by face using cosine metrics. Litter detection was done concurrently using YOLOv4, and the intersection area determined littering. To optimise system performance, only litterers were labelled.

In conclusion, the experimental setup encompassed several stages, including YOLOv4 fine-tuning, integration of the UKF with Deep Sort for person tracking, and face recognition using identification cards. By following this setup, the researchers aimed to improve litter detection accuracy, capture non-linear dynamics in person tracking, and enhance system robustness by replicating real-life scenarios.
\section{Results \& Discussions }

The first section of the model detects littering using a detection model and basic physics and geometry. Various object recognition models, such as YOLO, EfficientDet, Adaptive Training Sample Selection, Adaptive Spatial Feature Fusion, and CenterMask, are accessible as basis frameworks. Therefore, a comparative analysis is crucial to determine the most effective model for our use case. The selection was made using performance criteria such as Average Precision, Average Precision at 50\% IOU threshold, and Frames Per Second on a standard data set.

Average Precision (AP). Commonly used to test object detection, instance segmentation, and tracking model detection performance is AP. AP calculates the accuracy-recall curve area and detection output recall and precision. Low numbers indicate poor AP detection, while large numbers indicate good detection. In the graph, the Y axis indicates a percentage. FPS, or Frames Per Second. Tracking model computational efficiency is measured by FPS. It measures the tracking model's average frame rate per second. Tracking models with greater FPS are more efficient.

AP50, a variation of AP, assesses Average Precision at a 50\% IoU threshold. We employ this metric due to its effectiveness (\cite{ap}). Object detector detection performance is commonly assessed using this statistic. Due to its stricter assessment metric, AP50 requires the detection bounding box to coincide significantly with the ground-truth bounding box.

\begin{figure}[h]
\centering

\includegraphics[width=10cm]{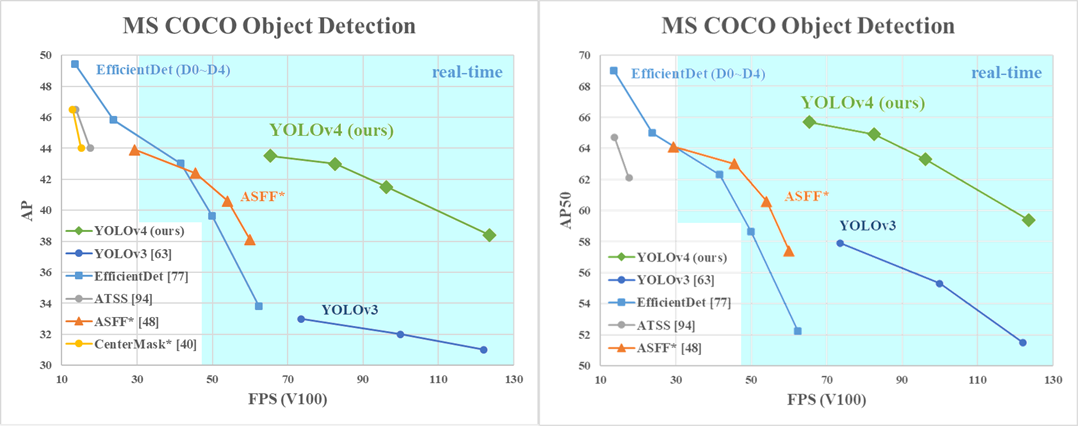}
\caption{For various object identification methods, the trade-off between accuracy (AP \& AP50) and computing efficiency (FPS).} \label{fig1}
\end{figure}

Figure 4 shows that YOLOv4 had the best trade-off. YOLO has the highest FPS, as demonstrated by \cite{perf}, which evaluates computing efficiency. This makes it the fastest model, which is crucial for speedy detection. YOLOv4's stronger AP and AP50 score made it the apparent pick because it delivered better precision and recall.
AP50 is more significant than AP for our issue statement since it addresses the case that will occur while running the model and detects littering action. The AP50 versus FPS graph shows that YOLOv4 has a slightly lower peak than EfficientDet but is far less erratic, reliable, and has a lot lower FPS. This enhanced our conviction that YOLOv4 is the optimal base detector model for our issue statement.
\begin{table}
\begin{center}
\caption{Comparison of Tracking Models}
\begin{tabular}{|c | c c c c c|} 
 \hline
 Model & HOTA & MOTA & IDF1 & AssA & DetA \\ [0.5ex] 
 \hline\hline
 SORT & 34.0 & 43.1 & 39.8 & 31.8 & 37.0 \\ 
 \hline
 DeepSORT & 61.2 & 78.0 & 74.5 & 59.7 & 63.1 \\
 \hline
 Improved & & & & &\\ DeepSORT & 63.6 & 78.6 & 78.5 & 62.3 & 63.6 \\
 \hline
\end{tabular}
\end{center}
\end{table}

We compared our model to DeepSORT and SORT using standardised parameters, as indicated in \cite{mota}, to evaluate the effectiveness of the enhanced DeepSORT tracking model when littering activity is detected. Table 1 shows that the model outperforms SORT and DeepSORT in practically all parameters.

\section{Conclusion}
The research successfully developed the Trashbuster system to automate the process of detecting and penalizing litterbugs. However, the study has limitations, such as detecting only twelve specific litter objects and a limited database for testing purposes. External factors like lighting, exposure, and camera features also impact the system's performance. Future improvements could include real-time implementation and collaboration with government agencies to promote accountability and raise awareness among offenders.

\end{document}